\documentclass[11pt]{article}
\usepackage{coling2020}
\usepackage{times}
\usepackage{url}
\usepackage{latexsym}

\usepackage{url}
\usepackage{epsfig}
\usepackage{latexsym}
\usepackage{microtype}
\usepackage{todonotes}
\usepackage{tabularx}
\usepackage{amsmath}
\usepackage{arydshln}
\usepackage{booktabs}
\usepackage{wrapfig}
\usepackage{textcomp}
\usepackage{chngpage}
\usepackage{graphicx,adjustbox}
\usepackage{multirow, makecell}
\usepackage[group-separator={,}, group-four-digits=true, group-digits=integer]{siunitx}
\usepackage[notext,not1]{stix}
\usepackage{fancyvrb}
\usepackage{tabularx}
\usepackage{colortbl}
\usepackage{mathtools}

\usepackage[T5,T1]{fontenc}

\usepackage{tikz}
\usepackage[linguistics]{forest}
\newcommand{\maybebf}[1]{{#1}}

\newcommand{\fig}[1]{Figure~\ref{fig:#1}}
\newcommand{\tab}[1]{Table~\ref{tab:#1}}

\newcommand{\hib}{\,$\uparrow$}
\newcommand{\lib}{\,$\downarrow$}

\colingfinalcopy 

\title{Aspects of Terminological and Named Entity Knowledge within Rule-Based Machine Translation Models for Under-Resourced Neural Machine Translation Scenarios}

\author{ Daniel~Torregrosa\(^1\) \qquad Nivranshu~Pasricha\(^1\) \qquad Maraim~Masoud\(^1\) \\ \qquad \textbf{Bharathi~Raja~Chakravarthi}\(^1\) \qquad \textbf{Juan~Alonso}\(^2\) \qquad \textbf{Noe~Casas}\(^2\) \qquad \textbf{Mihael~Arcan}\(^1\) \\
\(^1\) Insight Centre for Data Analytics, Data Science Institute, National University of Ireland Galway \\
\(^2\) United Language Group  \\
{\tt name.surname@insight-centre.org} \\ {\tt name.surname@ulgroup.com}
}

\begin{document}

\maketitle

\begin{abstract}
Rule-based machine translation is a machine translation paradigm where linguistic knowledge is encoded by an expert in the form of rules that translate text from source to target language. While this approach grants extensive control over the output of the system, the cost of formalising the needed linguistic knowledge is much higher than training a corpus-based system, where a machine learning approach is used to automatically learn to translate from examples. In this paper, we describe different approaches to leverage the information contained in rule-based machine translation systems to improve a corpus-based one, namely, a neural machine translation model, with a focus on a low-resource scenario. Three different kinds of information were used: morphological information, named entities and terminology. In addition to evaluating the general performance of the system, we systematically analysed the performance of the proposed approaches when dealing with the targeted phenomena. Our results suggest that the proposed models have limited ability to learn from external information, and most approaches do not significantly alter the results of the automatic evaluation, but our preliminary qualitative evaluation shows that in certain cases the hypothesis generated by our system exhibit favourable behaviour such as keeping the use of passive voice.
\end{abstract}

\section{Introduction}

In rule-based machine translation (RBMT), a linguist formalises linguistic knowledge into lexicons and grammar rules, which is used by the system to analyse sentences in the source language and translate them. While this approach does not require any parallel corpora for training and grants control over the translations created by the system, the process of encoding linguistic knowledge requires a great amount of expert time. Notable examples of RBMT systems are the original, rule-based Systran \cite{toma1977systran}, Lucy LT \cite{alonso2003comprendium} and the Apertium platform \cite{forcada2011apertium}.

Instead, corpus-based machine translation (MT) systems learn to translate from examples, usually in the form of sentence-level aligned corpora. On the one hand, this approach is generally computationally more expensive and offers limited control over the generated translations. Furthermore, it is not feasible for language pairs that have limited to no available parallel resources. On the other hand, if parallel resources are available, it boasts a much higher coverage of the targeted language pair. Examples of corpus-based MT paradigms are phrase-based statistical machine translation (PBSMT) \cite{koehn2003statistical} and neural machine translation (NMT) \cite{DBLP:journals/corr/BahdanauCB14}.

In this work, we focused on leveraging RBMT knowledge for improving the performance of NMT systems in an under-resourced scenario. Namely, we used the information provided by Lucy LT, an RBMT system where the linguistic knowledge is formalised by human linguists as computational grammars, monolingual and bilingual lexicons. Grammars are collections of transformations to annotated trees. Monolingual lexicons are collections of lexical entries, where each lexical entry is a set of feature-value pairs containing morphological, syntactic and semantic information. Bilingual lexicon entries include source-target lexical correspondences and, optionally, contextual conditions and actions. The Lucy LT system divides the translation process into three sequential phases: analysis, transfer, and generation. During the analysis phase, the source sentence is morphologically analysed using a lexicon that identifies each surface form and all its plausible morphological readings. Next, the Lucy LT chart parser together with an analysis grammar consisting of augmented syntactic rules extracts the underlying syntax tree structure and annotates it. The transfer and generation grammars are then applied in succession on that tree, which undergoes multiple annotations and transformations that add information about the equivalences in the target language and adapt the source language structures to the appropriate ones in the target language. Finally, the terminal nodes of the generation tree are assembled into the translated sentence. We focused on the analysis phase, with a special interest for two of the features used: the morphological category (\texttt{CAT}) and the inflexion class (\texttt{CL}) or classes of the lexical entries. 

Additionally, we focused on two language phenomena that are easily addressable when using RBMT but present a challenge when using corpus-based MT: named entities and terminological expressions.

A named entity (NE) is a word or a sequence of words that unequivocally refer to a real-world object, such as proper nouns, toponyms, numbers or dates. In the context of MT, NEs present different challenges. For example, if an English sentence starts with the word \textit{Smith}, we do not know a priori if we are dealing with the name of a profession, that will have to be translated, or a proper noun that may have to be left untranslated, or maybe transliterated to a different script. A second issue may arise when using subword units: while word-level models may accidentally preserve an out-of-vocabulary NE, the subword level model will generate a (most likely nonsensical) \textit{translation} for it. NEs are one of the main out-of-vocabulary word classes, which often cause translation problems that seriously affect the meaning of the sentence \cite{li2018neural}.

Similarly, a terminological expression can consist of a single word or a sequence of words that may have a different meaning depending on the context or domain they appear. Hence, the translation for the term might be different depending on the context or domain. Moreover, different contexts and domains may impose additional restrictions on the language used, such as different modes or the use of active or passive voice, and the presence of particular terminology may suggest that a translation is not acceptable even if the meaning of the source sentence is preserved. Accurate terminology translation is crucial to produce adequate translations \cite{arcan2017leveraging}.

In this work we extend and further analyse the injection of morphological information technique that we proposed in a previous word \cite{torregrosa_mtsummit2019} and we propose an approach to NEs and terminology that does not rely on any particular technology and can be applied to any MT approach using any kind of resource to detect and translate the NEs and terminological expressions. 
To test our proposed approach, we focused on English-Spanish (both generic and medical domain), English-Basque, English-Irish and English-Simplified Chinese language pairs in an under-resourced scenario, using corpora with around one million parallel entries per language pair and domain. Additional test sets that contain several examples of terms, NEs and rich morphology have also been selected and used to further explore the performance of the proposed approaches. Results suggest that, while obtaining results that are not statistically significantly different than the baseline in several scenarios, the proposed approaches show appropriate behaviours such as keeping the passive voice characteristic of some domains.


\section{Related Work}
In this section, we present the existing work on incorporating linguistic, terminological and NE information into NMT systems. 

\subsection{Use of Linguistic Knowledge}

Several approaches have been proposed to incorporate linguistic knowledge into MT models in order to improve translation quality. One of the approaches is to include the knowledge as features or extra tokens for the model. For example, morphological features, part of speech (POS) tags and syntactic dependency labels \cite{sennrich-haddow:2016:WMT} were proven to improve translation quality when translating between English and German and English to Romanian. A different approach used interleaved CCG supertags within the target word sequence \cite{nadejde2017}, comparing favourably to multi-task learning when translating from German and Romanian to English. Information can also be added to the target side by replacing it with a linearised and lexicalised constituency tree \cite{aharoni2017}, which shows improved word reordering when translating from German, Czech and Russian to English both in automatic and small-scale human evaluation.

A second approach is to modify the architecture of the recurrent neural network to capture linguistic knowledge. The encoder of the NMT ensemble was replaced with a graph convolutional network, that places no rigid constraints on the structure of the sentence \cite{bastings-etal-2017-graph}, which showed improvements when using syntactic dependency trees for the source language translating from English to German and Czech. An alternative approach modified the encoder to process tree-based syntactic representations of the source language, and the attention to be able to address both sentences and phrases \cite{eriguchi2016}, which improved results for English to Japanese translation.

A different approach is to use multi-task learning to improve translation quality by adding information from similar tasks, such as POS tagging. For example, two decoders were used to predict lemmas and factors (POS, gender, number, tense, person) independently \cite{GarcaMartnez2016FactoredNM} when translating from English to French, which led to increased vocabulary coverage. Another approach generated both the translation of the sentence, tagged the POS of the source sentence, and recognised NEs in the source language \cite{niehues-cho-2017-exploiting}. Different architectures that shared encoders, attention mechanisms and even decoders were used, showing improvements of all individual tasks when translating from German to English

Finally, different subword unit strategies have been tested. Generating compositional representations of the input words by using an auxiliary recurrent neural network \cite{ataman2018compositional} showed improved results compared to systems using byte-pair encoding when translating from morphologically rich languages (Arabic, Czech, German, Italian and Turkish) to English. Another alternative used morpheme-based segmentation \cite{banerjee-bhattacharyya-2018-meaningless}, which compared favourably to byte-pair encoding when translating English to Hindi, English to Bengali and Bengali to Hindi; what is more, a combination of both strategies showed even better results. Other representations, such as linguistically motivated or frequency-based word segmentation methods \cite{etchegoyhen2018}, were also explored when using NMT, RBMT and PBSMT.

It has also been investigated whether the encoder of NMT models learns syntactic information from the source sentence \cite{shi-etal-2016-string} when performing three different tasks: translating from English to French and English to German, generating a linearised constitutional tree from English, and auto-encoding from English to permutated English. The authors found that different types of syntactic information are captured in different layers.

\subsection{Terminology and Named Entities}

Several strategies have been tested for dealing with NE translation. 
For example, identifying NEs before translating and replacing the tokens with special tags or translating the NE using an external translation model \cite{yan2018impact}. This model showed performance improvements over the baseline model when translating sentences with person names from Simplified Chinese to English. 
A different approach used alignment information to align source and target language NEs before translating \cite{li2018neural}. As using information from both sides can help improving NE tagging, the model showed improvements over the baseline when translating from Simplified Chinese to English. 
Addressing multi-word NEs by using additional features to indicate where each NE starts and ends was also investigated \cite{ugawa2018neural}, which showed improvements when translating from English to Japanese, Romanian and Bulgarian.

Similarly, terminology translation has been approached in different ways.
The use of a cache-based model within PBSMT capable of combining both an static phrase table and language model with smaller, dynamic-loaded extensions \cite{arcan2014enhancing,arcan2017leveraging} compared favourably both to the baseline model and an XML-based markup mode that enables enforcing the translation of some tokens in the sentence (i.e.\ enforcing a particular translation for a term) when translating between English and Italian and English and German. 
A mechanism named guided NMT decoding \cite{chatterjee2017guiding}, similar in concept to the XML-based markup for PBSMT, was also tested, comparing favourably to baseline models, both in English to German translation and automatic post-editing.\footnote{ That is, \emph{translating} from English that is likely to have low adequacy, usually MT hypotheses, to post-edited, more adequate English.} This model was only able to guide the decoder, but not to enforce the restrictions; hence, a multi-stack approach using finite-state acceptor to enforce the constraints was proposed \cite{hasler2018neural}, showing improved results when translating from English to German and Simplified Chinese in scenarios using gold tokens and phrases present in the reference but not produced by the baseline system, or dictionaries. This information would be present in translation memories and glossaries provided by a possible customer. 
Finally, an approach that encodes the information encoded in knowledge graphs, i.e.\ terminological expressions and NEs, as embeddings that are then concatenated to the word embeddings was tested \cite{moussallem2019augmenting}, showing improved results for English to German translation.
Additionally, the performance of SMT and NMT have been explored when translating terminology without context, both using baseline and domain adapted models \cite{arcan2019translating}, showing that BPE-based NMT models benefit the most from domain adaptation.


\subsection{Data Selection}

Finally, data selection has been used to improve the performance of the trained models, reduce the computational cost of training, or both \cite{Rousseau13,chen2016bilingual}. Thought, to the best of our knowledge, applying data selection to the selection of targeted tests sets that frequently exhibit the studied feature to attain a higher insight of the performance of the model has not been previously explored in the literature.

\section{Methodology}


In this section, we describe the methodology to leverage rule-based machine translation (RBMT) information in neural machine translation (NMT).

\subsection{Information Acquisition From RBMT}

\begin{figure*}[t]
    \centering
        \begin{Verbatim}[fontsize=\small]
("snake" NST ALO "snake" CL (P-S S-01) KN CNT ON CO SX (N) TYN (ANI))
("snake" VST ALO "snak" ARGS ((($SUBJ N1 (TYN CNC LOC C-POT)) ($ADV DIR))) 
    CL (G-ING I-E P-ED PA-ED PR-ES1) ON CO PLC (NF))  
        \end{Verbatim}
    \caption{The word \textit{snake} as a noun (NST) and a verb (VST) in Lucy LT dictionaries. Each entry is composed of a canonical form, the category (POS), and a list of key-value features, such as the inflexion class (CL), the vocalic onset (ON), etc. }
    \label{fig:lucy_example}
\end{figure*}

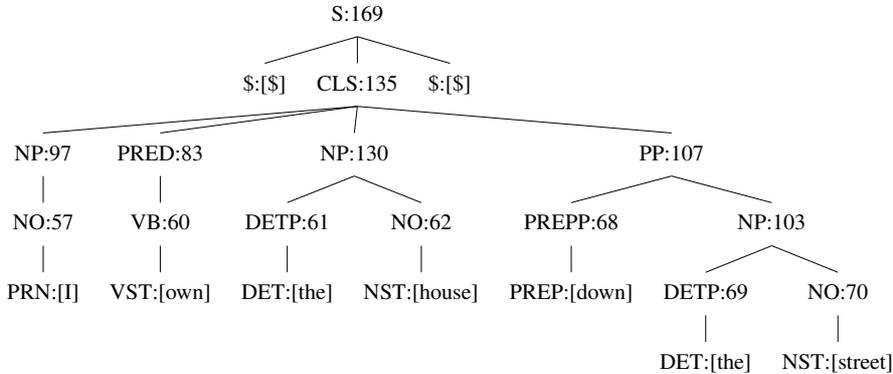
\begin{figure*}[t]
    \centering
    \adjustbox{trim=0 2ex 0 2ex}{\resizebox{.75\linewidth}{!}{
    \begin{forest}
[{{S:169}} [{{\$:}}{[{{\$}}]}] 
      [{{CLS:135}} [\maybebf{{NP}}{{:97}} [{{NO:57}} [\maybebf{{PRN}}:{[{{I}}]}]]] 
            [\maybebf{{PRED}}{{:83}} [{{VB:60}} [\maybebf{{VST}}:{[{{own}}]}]]] 
            [\maybebf{{NP}}{{:130}} [{{DETP:61}} [\maybebf{{DET}}:{[{{the}}]}]]
            [{{NO:62}} [\maybebf{{NST}}:{[{{house}}]}]]
                  ]
            [\maybebf{{PP}}:{{107}} [{{PREPP:68}} [\maybebf{{PREP}}:{[{{down}}]}]]
            [\maybebf{{NP}}{{:103}} [{{DETP:69}} [\maybebf{{DET}}:{[{{the}}]}]]
            [{{NO}}{{:70}} [\maybebf{{NST}}:{[{{street}}]}] ]]]] 
       [{{\$:{[\$]}}}]]
\end{forest}
}}
    \caption{Example of the parse tree for the English sentence \textit{I own the house down the street}.}
    \label{fig:tree}
\end{figure*}

Lucy LT monolingual lexicons are language-pair independent (i.e.\ the same English knowledge is used for all translation pairs including English as a source or target language) and mainly encode morphological and contextual information. Each entry has a word or multi-word expression (MWE) along with several features, such as the part of speech (POS) and morphological features. The bilingual lexicons mainly encode word-to-word or MWE-to-MWE translations and describe which target language word should replace each source language word. 
Still, the direct usage of the lexicon entries as a source of information presented a challenge, as there is no means to determine ambiguous surface words. For example, in English, most nouns will also be classified as verbs, as they share the same surface form; e.g.\ the word \textit{snake} can be both a noun and a verb (\fig{lucy_example}). For addressing this problem, we compare two different approaches: using \textbf{ambiguity classes} that describe all the possible analysis for a given surface word; and using \textbf{external information} (in the form of a monolingual POS tagger) for disambiguating ambiguous POS classes. 
For the former approach, we used a unique tag for each possible category (\texttt{CAT}) and class (\texttt{CL}) values concatenation. In the previous example, snake is both noun (\texttt{NST}) and verb (\texttt{VST}) (\fig{lucy_example}), so the value for the \texttt{CAT} feature would be \texttt{NST\_VST}.
For the latter, we used the Stanford POS tagger~\cite{toutanova2003feature}, that uses the Penn Treebank~\cite{Marcus:1994:PTA:1075812.1075835} tag set for English, the AnCora~\cite{civit2004building} tag set for Spanish. The IXA pipeline POS tagger~\cite{AGERRI14.775} with the Universal Dependencies POS tag set \cite{11234/1-2895} was used for the Basque language. All POS tag sets were mapped to the tag set used by Lucy LT. If the tagger provided POS tag was equivalent to one or more Lucy LT tags, then the non-matching Lucy LT tags were removed. Otherwise, we kept the set of tags; e.g.\ if the POS tag emits noun as the most likely tag, then only \texttt{NST} and the concatenation of all the inflexion classes for the corresponding entry would be used as additional information. As a comparison, we also evaluated NMT models trained with Stanford or IXA POS tags as additional information.

\subsection{Leveraging Syntactic Tree Information}

In addition to the direct use of the linguistic knowledge for the lexicon entries, the grammars (monolingual and bilingual lexicons) were indirectly used by exploring the results of each internal intermediate stage of the translation process, which Lucy LT expresses as annotated trees. 
For example, the sentence parsed in \fig{tree},
\begin{adjustwidth}{0.25in}{0.25in}
\textit{I own the house down the street}
\end{adjustwidth}
is encoded as
\begin{adjustwidth}{0.25in}{0.25in}
$\lParen$\textit{I own} $\lParen$\textit{the house}$\rParen$ $\lParen$\textit{down }$\lParen$\textit{the street}$\rParen\rParen\rParen$.\footnote{To avoid collisions with parenthesis in the text, we used the left ($\lParen$,~U+2985) and right ($\rParen$,~U+2986) white parenthesis.}
\end{adjustwidth}
We use this representation as source text when training the NMT models, as sequence-to-sequence deep neural network models do not generally accept hierarchical information. We also used an additional feature: the linguistic phrase the word belongs to. This information is present in the grandparent of each node; e.g.\ in \fig{tree} the noun \textit{house} appears in a noun phrase (NP).

\subsection{Named Entities and Terminology}

One of the main features of RBMT is that the linguist who is encoding the knowledge usually has full control over the output, letting the user define entries with more complex contexts that ensure that a certain possible translation case is covered. Conversely, corpus-based MT does not offer this feature: while the sentences used to train the system will have an impact on the words used when translating, it is not readily possible to enforce lexical selections. We devised two different strategies to address this situation.

The first strategy involved tagging each token with a feature that contains information about the kind of NE that the token belongs to, if any. Two different tag sets were used: a binary tag indicating if the token is part of a NE, and a collection of tags with the actual category of NE, according to CoreNLP classes. The second strategy involved replacing NEs with a special token. Like in the previous approach, we replaced each NE either with a generic token (similar to the binary tag) or with a special token representing the category of the NE. We used the same approach for terms, but as we only target the medical domain, there would be no difference between using binary tags or the actual classes. 

For example, for tagging medical (\textit{MED}) terminology, given the sentence 

\begin{adjustwidth}{0.25in}{0.25in}
\textit{He should discuss it with his cardiologist.}
\end{adjustwidth}

In the first approach, all words would get tagged with the domain as a feature:

\begin{adjustwidth}{0.25in}{0.25in}
{\textit{He}\texttt{|GEN} \textit{should}\texttt{|GEN} \textit{discuss}\texttt{|GEN} \textit{it}\texttt{|GEN} \textit{with}\texttt{|GEN} \textit{his}\texttt{|GEN} \textit{cardiologist}\texttt{|MED} \textit{.}\texttt{|GEN}}
\end{adjustwidth}

In the second approach, NEs or terms get replaced with the kind of NE or domain of the term:

\begin{adjustwidth}{0.25in}{0.25in}
{\textit{He should discuss it with his} \texttt{MED}\textit{.}}
\end{adjustwidth}

In the case dealing with NEs, \textit{cardiologist} would be tagged as \texttt{NE} when using binary tags and \textit{TITLE} when using CoreNLP classes.

Two different approaches were taken when detecting which tokens are part of a NE or term. During training, we detected NEs in the source side using CoreNLP, aligned source and target using \texttt{eflomal} \cite{Ostling2016efmaral}, and replaced the source tokens and the corresponding aligned target tokens with the source tag. In the case of terminology, we used the information contained in Lucy LT bilingual lexicons. When translating, we detected NEs and terms in the source language using CoreNLP or Lucy LT information and replaced or tagged them with the corresponding labels.

We used the same pre-processing both when training the NMT models and when translating.
After translating, each tag generated in the hypothesis sentence is aligned to the most likely tag in the source sentence using the soft-alignment produced by the attention mechanism, and replaced with the actual translation of the NE or terminological expression.


To obtain the actual translation for NEs and terms, we used the Lucy LT lexicons, selecting the entry corresponding to the targeted domain in the case of terminology translation. As a comparison, we used Google Translate to generate translations for the NEs and terms; while unfair due to the lack of context and the lack of an option to select a specific domain for the translation, it can be used as a baseline for our method. Additionally, the OpenNMT feature that lets user include a phrase table to replace unknown tokens was used; as it can only handle one-token phrases, the dictionary extracted from Lucy LT was aligned using \texttt{eflomal} and the most likely alignment for each word was used as a phrase.

In some cases, a sentence may have two or more NEs or terms of the same kind. When using the replacement strategy, it is possible that the model will not learn how to properly align the tags with the correct source words. For this reason, we have also tested a system where sentences that have NEs or terms have been duplicated and kept intact, i.e. as if the NEs or terms were generic words instead. This approach slightly increases the size of the training set, but does not add new information to the training approach.

Finally, in the case of terminology translation, we compared the performance of our approach against back-translation, a commonly taken approach in scenarios where domain adaptation is needed \cite{sennrich2015improving}.

On the one hand, this approach is completely independent regarding the MT implementation and the resource used to detect and/or obtain translations for NEs and terminological expressions, hence being applicable in many scenarios. On the other hand, some MT implementations (namely, most corpus-based MT ones) will not guarantee that the generated hypothesis will contain all the tags present in the source sentence, which can lead to lower translation adequacy.

\subsection{Focused Evaluation}

In this work, we targeted several phenomena that appear when translating a sentence, namely morphology, NEs and terminology. While the quantitative or qualitative evaluation of the models using the same test sets is necessary, it might not properly capture the improvements to the targeted phenomena. For this reason, we proposed an additional evaluation focused on the targeted phenomena by using specially selected corpora. In the case of morphology, we selected a corpus that contained the Spanish verbs \textit{tener} (to have), \textit{poder} (can/may) and \textit{decir} (to say) in different surface forms. In the case of NEs and terminology, we selected sentences that contained NEs or terminology according to CoreNLP or Lucy LT respectively. All these sentences do not appear in any training or development set.

\section{Experimental Setting}

In this section, we describe the resources we used to train and evaluate the systems, along with the NMT framework used.


\subsection{Training and Evaluation Datasets}

\def\tlabel{t}
\def\vlabel{v}
\def\elabel{e}
\def\nwords{\(|\mathrm{Words}|\)}
\def\nuwords{\(|\mathrm{Vocab}|\)}
\def\nbpe{\(|\mathrm{Subwords}|\)}
\def\nubpe{\(|\mathrm{Vocab}_s|\)}
\def\numlin{\(|\mathrm{Lines}|\)}

\begin{table}[th!]
    \centering
    \small
    \resizebox{\textwidth}{!}{
        \begin{tabular}{c@{\hspace{1ex}}c@{\hspace{2ex}}S[table-format=8.0]S[table-format=5.0]S[table-format=8.0]S[table-format=5.0]S[table-format=8.0]S[table-format=5.0]S[table-format=8.0]S[table-format=5.0]S[table-format=7.0]}
\toprule
 &   &  \multicolumn{4}{c}{Source (English)}  &  \multicolumn{4}{c}{Target}  \\ 
\midrule
 &   & {\nwords}  & {\nuwords}  &  {\nbpe}  &  {\nubpe}  &  {\nwords}  & {\nuwords}  &  {\nbpe}  &  {\nubpe}  &  {\numlin}  \\
\midrule
\midrule
\multirowcell{3}{English--\\--Spanish\\(generic)}
 &  \tlabel{}  & 15495771 & 253973 & 17919926 & 33212 & 15863310 & 300015 & 18408749 & 33076 & 991880 \\
 &  \vlabel{}  & 155264 & 16999 & 180290 & 15714 & 159214 & 21021 & 185662 & 18804 & 9917 \\
 &  \elabel{}  & 154666 & 16649 & 178841 & 15031 & 157450 & 20762 & 181188 & 18810 & 9921 \\
\midrule
\multirowcell{3}{English--\\--Spanish\\(EMEA)} 
 &  \tlabel{}  & 13180375 & 57990 & 14165448 & 23363 & 14660270 & 70547 & 15533465 & 26872 & 1032842 \\
 &  \vlabel{}  & 165577 & 12913 & 179581 & 11387 & 183697 & 15239 & 196614 & 13930 & 9931 \\
 &  \elabel{}  & 163104 & 12824 & 208848 & 8995 & 182693 & 15208 & 233219 & 10526 & 9936 \\
\midrule
\multirowcell{3}{English--\\--Basque}   
 &  \tlabel{}  & 10766339 & 115978 & 11760808 & 30946 & 8699001 & 246552 & 10309229 & 32369 & 1357475 \\
 &  \vlabel{}  & 78077 & 9697 & 85919 & 9150 & 63607 & 14894 & 76532 & 13593 & 10000 \\
 &  \elabel{}  & 77655 & 9608 & 85163 & 9283 & 63273 & 14564 & 75309 & 13546 & 10000 \\
\midrule 
\multirowcell{3}{English--\\--Irish}  
 &  \tlabel{}  & 14854747 & 133808 & 15234432 & 31834 & 16058640 & 229516 & 16983046 & 32183 & 1090418 \\
 &  \vlabel{}  & 132178 & 12564 & 135986 & 12648 & 143608 & 17039 & 152224 & 16113 & 10000 \\
 &  \elabel{}  & 133314 & 12342 & 140696 & 11613 & 143046 & 16959 & 152064 & 16174 & 10000 \\
\midrule
\multirowcell{3}{English--\\--Simplified\\Chinese} 
 &  \tlabel{}  & 26185343 & 258534 & 27878268 & 31471 & 23784689 & 301656 & 25199106 & 41458 & 995000 \\
 &  \vlabel{}  & 130361 & 13313 & 138640 & 12451 & 119277 & 14939 & 126191 & 14490 & 5000 \\
 &  \elabel{}  & 121309 & 13017 & 129440 & 12175 & 111691 & 14746 & 119577 & 14431 & 4500 \\
\bottomrule
    \end{tabular}
    }
    \caption{Statistics on the used train (\tlabel), validation (\vlabel) and evaluation (\elabel) datasets. English-Spanish (EMEA) is a subset of the whole EMEA corpus.}
    \label{tab:data_stats}
\end{table}

\begin{table}[th!]
    \centering
    \small
    \resizebox{\textwidth}{!}{
        \begin{tabular}{c@{\hspace{1ex}}c@{\hspace{2ex}}S[table-format=8.0]S[table-format=5.0]S[table-format=8.0]S[table-format=5.0]S[table-format=8.0]S[table-format=5.0]S[table-format=8.0]S[table-format=5.0]S[table-format=7.0]}
\toprule
 &   &  \multicolumn{4}{c}{English}  &  \multicolumn{4}{c}{Spanish}  \\ 
\midrule
 &   & {\nwords}  & {\nuwords}  &  {\nbpe}  &  {\nubpe}  &  {\nwords}  & {\nuwords}  &  {\nbpe}  &  {\nubpe}  &  {\numlin}  \\
\midrule
\midrule
\multirowcell{3}{EMEA}
 &  \tlabel{}  & 2621489 & 35786 & 3320354 & 14667 & 2920259 & 43084 & 3691632 & 17283 & 200000 \\
 &  \vlabel{}  & 36817 & 5961 & 47569 & 5431 & 39887 & 6642 & 47569 & 6024 & 2000 \\
\midrule
\multirowcell{1}{Word BT$_1$}  &  \tlabel{}  & 2693900 & 38384 & {--} & {--} & 2759072 & 29081 & {--} & {--} & 200000 \\
\multirowcell{1}{BPE BT$_1$}  &  \tlabel{}  & {--} & {--} & 3271935 & 15045 & {--} & {--} & 3523079 & 16913 & 200000 \\
\midrule
\multirowcell{1}{Word BT-$_2$}  &  \tlabel{}  & 3085550 & 14137 & {--} & {--} & 2601687 & 16128 & {--} & {--} & 200000 \\
\multirowcell{1}{BPE BT-$_2$}  &  \tlabel{}  & {--} & {--} & 3357112 & 41112 & {--} & {--} & 3563324 & 19251 & 200000 \\
\midrule
\multirowcell{1}{BPE BT-$_3$} &  \tlabel{}  & {--} & {--} & 3339520 & 14110 & {--} & {--} & 3555254 & 16225 & 200000 \\
 \midrule
\multirowcell{2}{Baseline+} 
 &  \tlabel{}  & 15766197 & 255145 & 17195781 & 32360 & 16169821 & 301174 & 17608787 & 33812 & 1001639 \\
 &  \vlabel{}  & 157836 & 17060 & 172818 & 15200 & 162036 & 21084 & 176928 & 18299 & 10000 \\
\bottomrule
    \end{tabular}
    }
    \caption{Statistics on the used train (\tlabel), validation (\vlabel) and evaluation (\elabel) datasets for the models focused on terminology translation. EMEA contains the statistics for the subset of the EMEA corpus that was back-translated (BT), and BT$_i$ contains the stats for the $i$-th round of back-translation with word-based or BPE-based models. The same validation set was used for all BT models. Baseline+ duplicates those lines with detected medical terms, going from \num{991880} to \num{1001639}, an increase of \num{9759} lines.}
    \label{tab:data_stats_term}
\end{table}

\begin{table}[th!]
    \centering
    \small
    \resizebox{\textwidth}{!}{
        \begin{tabular}{c@{\hspace{1ex}}c@{\hspace{2ex}}S[table-format=8.0]S[table-format=5.0]S[table-format=8.0]S[table-format=5.0]S[table-format=8.0]S[table-format=5.0]S[table-format=8.0]S[table-format=5.0]S[table-format=7.0]}
\toprule
 & & \multicolumn{4}{c}{English}  &  \multicolumn{4}{c}{Spanish}  \\ 
\midrule
 & & {\nwords} & {\nuwords}   &  {\nbpe} &  {\nubpe}  &  {\nwords}  & {\nuwords}  & {\nbpe} &  {\nubpe}  &  {\numlin}  \\
\midrule
\midrule
{Named Entities} & \elabel{} & 85339  & 9452 & 95587 & 9247 &  91815  & 10704 & 111403 & 9781 & 3031 \\
{Morphology} & \elabel{} & 22923  & 4627 & 24765 & 4917& 24123  & 5365 & 27729 & 5534 &  1020 \\
{Terminology} &  \elabel{}  & 163104 & 12824 & 208848 & 8995 & 182693 & 15208 & 233219 & 10526 & 9936 \\
\bottomrule
\end{tabular}
    }
    \caption{Statistics on the named entities, terminology and morphology focused evaluation (\elabel) datasets. }
    \label{tab:ne_data_stats}
\end{table}

In this work, besides studying the impact of leveraging RBMT knowledge into NMT systems, we further focused on NMT for under-resourced scenarios. On the one hand, we consider languages, such as Basque or Irish, which do not have a significant amount of parallel data necessary to train a neural model. On the other hand, an under-resourced scenario can be a specific domain, e.g.\ medical, where a significant amount of data exists, but does not cover the targeted domain. The Table \ref{tab:data_stats} shows the statistics on the used datasets.

For English-Basque and English-Irish, we used the available corpora stored on the OPUS webpage.\footnote{\url{opus.nlpl.eu}} We used OpenSubtitles2018~\cite{LISON16.947},\footnote{\url{opensubtitles.org}} Gnome and KDE4 datasets~\cite{Tiedemann:2012}. Additionally, the English-Irish parallel corpus is augmented with second level education textbooks (\textit{Cuimhne na dTéacsleabhar}) in the domain of economics and geography \cite{ARCAN16.9}.

In addition to that, we also focused on well resourced languages (Spanish and Simplified Chinese), but limited the training datasets to around one million aligned sentences. To ensure a broad lexical and domain coverage of our NMT system, we merged the existing English-Spanish parallel corpora from the OPUS web page into one parallel data set and randomly extracted the sentences. In addition to the previous corpora, we added Europarl~\cite{Koehn:2005}, DGT~\cite{DBLP:journals/lre/SteinbergerEPCSPG14}, MultiUN corpus \cite{MultiUN}, EMEA and OpenOffice~\cite{Tiedemann2009}. 
Sentences extracted from the rest of the corpus were used for the targeted evaluation.
To evaluate the targeted under-resourced scenario within medical domain and terminology translation, we exclusively used the EMEA corpus.

For Simplified Chinese, we used a parallel corpus provided by the industry partner, which was collected from bilingual English-Simplified Chinese news portals.
The corpora were tokenised using the OpenNMT toolkit and lowercased, with the exception of Simplified Chinese, that was tokenized using Jieba.\footnote{\url{github.com/fxsjy/jieba}}

Some experiments used or generated additional data; namely, those evaluating the different strategies on specific corpus exhibiting the studied feature. Statistics for those corpora are described in \tab{data_stats_term} and \tab{ne_data_stats}.


\subsection{NMT Framework}

We used OpenNMT~\cite{2017opennmt}, a generic deep learning framework mainly specialised in sequence-to-sequence models covering a variety of tasks such as machine translation, summarisation, speech processing and question answering as NMT framework. Due to computational complexity, the vocabulary in NMT models had to be limited. To overcome this limitation, we used byte pair encoding (BPE) to generate subword units \cite{journals/corr/SennrichHB15}. BPE is a form of data compression that iteratively replaces the most frequent pair of bytes in a sequence with a single, unused byte. We also added the different morphological and syntactic information as word features. 

We used the following default neural network training parameters: two hidden layers, \num{500} hidden LSTM (long short term memory) units per layer, input feeding enabled, \num{13} epochs, batch size of \num{64}, \num{0.3} dropout probability, dynamic learning rate decay, \num{500} dimension embeddings, unlimited different values for the word features and between \num{11} and \num{23} dimension embeddings for word features.\footnote{The size of the embedding for word features depend on the number of unique values for the feature.}  For word models, we used a maximum vocabulary size of \num{50000} words. For subword models, we used or subwords, a maximum vocabulary size of \num{50000} and a maximum of \num{32000} unique BPE merge operations.

\subsection{Evaluation Metrics} 
In order to evaluate the performance of the different systems, we used BLEU~\cite{papineni2002bleu}, an automatic evaluation that boasts high correlation with human judgements, and translation error rate (TER)~\cite{snover2006study}, a metric that represents the cost of editing the output of the MT systems to match the reference, and \textbf{chrF3}~\cite{popovic:2015:WMT}, a character n-gram metric which shows very good correlations with human judgements on the WMT2015 shared metric task~\cite{stanojevic-EtAl:2015:WMT}, especially when translating from English into morphologically rich(er) languages.


Additionally, we used bootstrap resampling~\cite{koehn2004statistical} with a sample size of \num{1000} and \num{1000} iterations, and reported statistical significance with $p<0.05$. 
In addition, we compared the performance of our NMT systems with the NMT-based Google Translate,\footnote{\url{translate.google.com/} retrieved between March and August 2019.}
and the translations performed using Lucy LT RBMT; for the latter, only English-Spanish and English-Basque models are available.

\section{Results}
\label{sec:eval}

In this section, we describe the quantitative and qualitative evaluation of the different models: the NMT baseline (Baseline), baseline enhanced with ambiguous \texttt{CAT} and \texttt{CL} (CAT-CL), baseline with disambiguated \texttt{CAT} and \texttt{CL} (CAT-CL D), baseline with external POS tags (POS), baseline with indirect \texttt{CAT}, \texttt{CL} and syntactic information (CAT-CL L), the hierarchical model (Tree), Lucy LT (RBMT) and Google Translate (Google). Systems that are not shared between evaluations are described in the corresponding subsection.

\subsection{Quantitative Results}

In this section, we describe the quantitative evaluation of the different models.

\subsubsection{General Evaluation}

\begin{table}[th!]
    \small
    \centering
    \begin{tabular}{rr@{\hskip 3\tabcolsep}SSS[table-format=0.4]@{\hskip 3\tabcolsep}SSS[table-format=0.4]}
		\multicolumn{2}{r@{\hskip 3\tabcolsep}}{English $\rightarrow$} & \multicolumn{3}{c@{\hskip 3\tabcolsep}}{Spanish} & \multicolumn{3}{c}{Basque}\\ 
		\multicolumn{2}{r@{\hskip 3\tabcolsep}}{} & {BLEU\hib}  & {chrF\hib} & {TER\lib} & {BLEU\hib} & {chrF\hib} & {TER\lib} \\ 
		\midrule
		\multirow{6}{*}{\rotatebox[origin=c]{90}{\centering Word}} 
        & Baseline & 34.22 & 58.87 & 0.5128 & 30.78 & 59.41 & 0.5497 \\
		& CAT{-}CL & 34.46 & 58.99 & 0.5107 & 31.54 & 60.05 & 0.5433 \\
		& CAT{-}CL D & 34.19 & 58.9 & 0.5096 & 31.5 & 59.99 & 0.5447 \\
		& CAT{-}CL L & 34.56 & 59. & 0.5104  & 30.6 & 60.14 & 0.5502 \\
		& Tree & 31.2 & 57.17 & 0.547 & 26.22 & 57.42 & 0.5887 \\
		& POS & 34.47 & 58.99 & 0.5101 & 31.81 & 59.72 & 0.5455 \\
		\cmidrule{2-8}
		\multirow{6}{*}{\rotatebox[origin=c]{90}{\centering BPE}} 
		& Baseline & 35.68 & 60.38 & 0.5074 & 32.95 & 63.23 & 0.5292 \\
		& CAT{-}CL & 36.14 & 60.64 & 0.5028 & 32.75 & 62.84 & 0.5353 \\
		& CAT{-}CL D & 35.67 & 60.41 & 0.5048 & 32.73 & 63.05 & 0.5346 \\
		& CAT{-}CL L & 36.08 & 60.51 & 0.5044 & 31.91 & 62.93 & 0.5381 \\
		& Tree & 32.94 & 59.17 & 0.5307 & 28.18 & 60.61 & 0.5654 \\
		& POS & 36.09 & 60.83 & 0.5012 & 32.82 & 62.84 & 0.537 \\
		\cmidrule{2-8}
		\multirow{2}{*}{\rotatebox[origin=c]{90}{\centering }} 
		& RBMT & 24.52 & 51.71 & 0.6058 & 11.09 & 41.27 & 0.8078 \\
		& Google & 38.69 & 61.28 & 0.4942 & 19.11 & 51. & 0.679 \\	
		\arrayrulecolor{white}\midrule\arrayrulecolor{black}   
		\multicolumn{2}{r@{\hskip 3\tabcolsep}}{English $\leftarrow$} & \multicolumn{3}{c@{\hskip 3\tabcolsep}}{Spanish} & \multicolumn{3}{c}{Basque}\\ 
		\multicolumn{2}{r@{\hskip 3\tabcolsep}}{} & {BLEU\hib}  & {chrF\hib} & {TER\lib} & {BLEU\hib} & {chrF\hib} & {TER\lib} \\ 
		\midrule
		\multirow{6}{*}{\rotatebox[origin=c]{90}{\centering Word}} 
		& Baseline & 34.59 & 58.29 & 0.499 & 33.27 & 57.34 & 0.5158 \\
		& CAT{-}CL & 34.68 & 58.15 & 0.4992 & 31.2 & 56.41 & 0.5328 \\
		& CAT{-}CL D & 34.79 & 58.5 & 0.4963 & 32. & 57.73 & 0.5171 \\
		& CAT{-}CL L & 34.63 & 58.48 & 0.4953 & 33.4 & 57.93 & 0.511 \\
		& Tree & 27.66 & 54.01 & 0.5636 & 28.32 & 56.13 & 0.5439 \\
		& POS & 34.68 & 58.07 & 0.5 & 26.49 & 53.84 & 0.5675 \\
		\cmidrule{2-8}
		\multirow{6}{*}{\rotatebox[origin=c]{90}{\centering BPE}} 
		& Baseline & 36.26 & 60.55 & 0.4838 & 35.86 & 61.29 & 0.4859 \\
		& CAT{-}CL & 36.5 & 60.22 & 0.4898 & 32.42 & 59.44 & 0.5153 \\
		& CAT{-}CL D & 36.49 & 60.69 & 0.4887 & 29.12 & 58.85 & 0.5692 \\
		& CAT{-}CL L & 36.79 & 60.81 & 0.4853 & 35.15 & 60.84 & 0.4967 \\
		& Tree & 28.71 & 56.51 & 0.551 & 29.48 & 56.99 & 0.5391 \\
		& POS & 36.4 & 60.62 & 0.4845 & 29.42 & 59.53 & 0.5653 \\
		\cmidrule{2-8}
		\multirow{2}{*}{\rotatebox[origin=c]{90}{\centering }} 
		& RBMT & 23.27 & 49.41 & 0.6288 & 13.03 & 37.23 & 0.7982 \\
		& Google & 40.08 & 62.07 & 0.4711 & 26.23 & 52.3 & 0.5836 \\
    \end{tabular}
    \caption{Results for the evaluation for English-Spanish and English-Basque. Models marked with $\ast$ are significantly better than the NMT BPE-based baseline. All BPE models are statistically significantly better than their word-based counterparts. All models are statistically significantly better than RBMT, and all models for English-Basque and Basque-English are statistically significantly better than Google Translate.}
    \label{tab:results_es_eu}
\end{table}

\begin{table}[th!]
    \centering
    \small
    \begin{tabular}{rr@{\hskip 3\tabcolsep}SSS[table-format=0.4]@{\hskip 3\tabcolsep}SSS[table-format=0.4]}
        \multicolumn{2}{r@{\hskip 3\tabcolsep}}{English $\rightarrow$} & \multicolumn{3}{c@{\hskip 3\tabcolsep}}{Irish} & \multicolumn{3}{c}{Simplified Chinese}\\ 
        \multicolumn{2}{r@{\hskip 3\tabcolsep}}{} & {BLEU\hib}  & {chrF\hib} & {TER\lib} & {BLEU\hib} & {chrF\hib} & {TER\lib} \\ 
        \midrule
        \multirow{4}{*}{\rotatebox[origin=c]{90}{\centering Word}} 
        & Baseline & 49.48 & 72.4 & 0.4017 & 27.94 & 58.1 & 0.5538 \\
        & CAT-CL & 49.45 & 72.29 & 0.4053 & 27.67 & 57.76 & 0.5692 \\
        & CAT-CL D & 49.62 & 72.5 & 0.4049 & 27.55 & 58.27 & 0.5533 \\
        & POS & 50.11 & 72.53 & 0.4047 & 28.1 & 57.99 & 0.5551 \\
        \cmidrule{2-8}
        \multirow{4}{*}{\rotatebox[origin=c]{90}{\centering BPE}} 
        & Baseline & 50.11 & 72.8 & 0.4003 & 28.43 & 58.67 & 0.5509 \\
        & CAT-CL & 49.87 & 72.87 & 0.3983 & 28.72 & 58.57 & 0.5489 \\
        & CAT-CL D & 49.02 & 72.43 & 0.4072 & 28.49 & 58.67 & 0.5513 \\
        & POS & 49.99 & 72.37 & 0.4107 & 28.19 & 58.41 & 0.5573 \\
        \cmidrule{2-8}
        & Google & 38.97 & 66.37 & 0.4914 & 27.37 & 58.53 & 0.5452 \\
        \arrayrulecolor{white}\midrule\arrayrulecolor{black}   
        \multicolumn{2}{r@{\hskip 3\tabcolsep}}{English $\leftarrow$} & \multicolumn{3}{c@{\hskip 3\tabcolsep}}{Irish} & \multicolumn{3}{c}{Simplified Chinese}\\ 
        \multicolumn{2}{r@{\hskip 3\tabcolsep}}{} & {BLEU\hib}  & {chrF\hib} & {TER\lib} & {BLEU\hib} & {chrF\hib} & {TER\lib} \\ 
        \midrule
        & Word & 56.48 & 74.38 & 0.3462 & 34.02 & 61.39 & 0.5326 \\
        \cmidrule{2-8}
        & BPE & 57.16 & 75.9 & 0.3374 & 35.01 & 61.98 & 0.528 \\
        \cmidrule{2-8}
        & Google & 42.21 & 65.05 & 0.4587 & 37.41 & 62.89 & 0.5112 \\
    \end{tabular}
    \caption{Results for the evaluation for English-Irish and English-Simplified Chinese. All BPE models for English-Chinese, Chinese-English and Irish-English are statistically significantly better than their word-based counterparts. No RMBT models are available for Irish and Simplified Chinese in Lucy LT, and all models for English-Irish and Irish-English are statistically significantly better than Google Translate.}
    \label{tab:results_ga_zh}
\end{table}

The quantitative results of the evaluation are presented in \tab{results_es_eu} and \tab{results_ga_zh}. 
All the models tested significantly outperformed the RBMT system Lucy LT both when using BLEU and TER as evaluation metrics. 
Even when trained with only around a million sentences, the NMT baseline model for English-Basque and English-Irish performed better than Google Translate with generic domain corpora, and were not statistically significantly different for English$\rightarrow$Simplified Chinese. Conversely, Google Translate was significantly better than the NMT baselines only for the English-Spanish generic domain, excluding English$\rightarrow$Spanish TER. While some of the feature-enriched models obtained slightly better results in terms of BLEU and TER compared to the baseline, no model obtains scores that are statistically significantly different than the baseline subword model. In the case of the tree model, the results were consistently lower than the rest. We learned that the system could not cope with this complex representation with the amount of data available.


\subsubsection{Evaluation Focused on Morphological Information}

\begin{table}[th!]
    \centering
    \small
    \begin{tabular}{rr@{\hskip 3\tabcolsep}SSS[table-format=0.4]@{\hskip 3\tabcolsep}SSS[table-format=0.4]}
		\multicolumn{2}{r@{\hskip 3\tabcolsep}}{} & \multicolumn{3}{c@{\hskip 3\tabcolsep}}{English$\rightarrow$Spanish} & \multicolumn{3}{c}{English$\leftarrow$Spanish}\\ 
		\multicolumn{2}{r@{\hskip 3\tabcolsep}}{} & {BLEU\hib}  & {chrF\hib} & {TER\lib} & {BLEU\hib} & {chrF\hib} & {TER\lib} \\ 
		\midrule
		\multirow{6}{*}{\rotatebox[origin=c]{90}{\centering Word}} 
        & Baseline & 29.2 & 57.86 & 0.539 & 27.84 & 51.54 & 0.5782 \\
		& CAT{-}CL & 29.17 & 58.06 & 0.5352 & 28.4 & 51.83 & 0.5669 \\
		& CAT{-}CL D & 28.72 & 58.01 & 0.5365 & 28.29 & 51.69 & 0.5654 \\
		& CAT{-}CL L & 29.55 & 57.76 & 0.5376 & 27.61 & 51.3 & 0.5741 \\
		& Tree & 27.68 & 57.43 & 0.5613 & 23.61 & 48.27 & 0.629 \\
		& POS  & 28.72 & 57.86 & 0.5399 & 28.37 & 52.03 & 0.5636 \\
		\cmidrule{2-8}
		\multirow{6}{*}{\rotatebox[origin=c]{90}{\centering BPE}} 
		& Baseline & 29.7 & 59.04 & 0.5367 & 30.36 & 56.04 & 0.559 \\
		& CAT{-}CL & 30.8${\ast}$ & 59.39 & 0.5291 & 30.79 & 55.99 & 0.553 \\
		& CAT{-}CL D & 29.95 & 59.13 & 0.5328 & 30.6 & 56.22 & 0.5551 \\
		& CAT{-}CL L & 30.09 & 58.88 & 0.5353 & 30.83 & 56.15 & 0.5496 \\
		& Tree & 36.98${\ast}$ & 62.28${\ast}$ & 0.5099${\ast}$  & 35.64${\ast}$ & 58.72${\ast}$ & 0.531${\ast}$ \\
		& POS & 30.67${\ast}$ & 59.45 & 0.5297 & 30.38 & 56.13 & 0.5521 \\
		\cmidrule{2-8}
		\multirow{2}{*}{\rotatebox[origin=c]{90}{\centering }} 
		& RBMT & 19.3 & 50.12 & 0.6521 &   17.25 & 44.9 & 0.7056 \\
		& Google & 32.41 & 60.12 & 0.5161 &  36.59 & 60.26 & 0.4983  \\	
\end{tabular}
    \caption{Results for the generic models when tested with the morphology-focused test set. Models marked with $\ast$ are significantly better than the NMT Baseline model. All BPE models are statistically significantly better than their word-based counterparts, and all models are statistically significantly better than RBMT.}
    \label{tab:results_mo}
\end{table}

The results presented in \tab{results_mo} show that the added morphological information has a greater impact when using the BPE model, especially when translating from English to Spanish, that is, from a language with less morphological information to one with more. Also, in this scenario the tree-based model vastly outperforms the baseline when using subword units; the structural information along with the morphological information is helping the system make better decisions when translating this corpus, that has a high density of verbs.

\subsubsection{Evaluation Focused on Named Entities}

\begin{table}[th!]
    \centering
    \small
    \begin{tabular}{rrr@{\hskip 3\tabcolsep}SSS[table-format=0.4]@{\hskip 3\tabcolsep}SSS[table-format=0.4]}
		\multicolumn{3}{r@{\hskip 3\tabcolsep}}{} & \multicolumn{3}{c@{\hskip 3\tabcolsep}}{English$\rightarrow$Spanish} & \multicolumn{3}{c}{English$\leftarrow$Spanish}\\ 
		\multicolumn{3}{r@{\hskip 3\tabcolsep}}{} & {BLEU\hib}  & {chrF\hib} & {TER\lib} & {BLEU\hib} & {chrF\hib} & {TER\lib} \\ 
		\midrule
		\multirow{10}{*}{\rotatebox[origin=c]{90}{\centering Word}} 
			& & Baseline & 34.22 & 58.87 & 0.5128 & 34.59 & 58.29 & 0.499 \\
			\cmidrule{3-9}
			& \multirow{2}{*}{\rotatebox[origin=c]{90}{\centering F}} 
			&   Binary & 34.14 & 58.73 & 0.5141 & 34.36 & 58.26 & 0.5025 \\
			& & NE & 34.25 & 58.78 & 0.515 & 34.04 & 58.4 & 0.4979 \\
            \cmidrule{3-9}
			& \multirow{2}{*}{\rotatebox[origin=c]{90}{\centering PG}} 
			&   Binary & 32.03 & 58.78 & 0.5352 & 32.64 & 58.77 & 0.5268 \\ 
			& & NE & 32.54 & 59.09 & 0.5307 & 32.41 & 58.64 & 0.5305 \\
            \cmidrule{3-9}
			& \multirow{2}{*}{\rotatebox[origin=c]{90}{\centering PL}} 
			&   Binary & 31.82 & 58.49 & 0.5378 & 32.89 & 58.6 & 0.527 \\
			& & NE & 32.34 & 58.8 & 0.5334 & 32.61 & 58.44 & 0.5308 \\
		\cmidrule{2-9}
		\multirow{7}{*}{\rotatebox[origin=c]{90}{\centering BPE}} 
			& & Baseline & 35.68 & 60.38 & 0.5074 & 36.26 & 60.55 & 0.4838 \\
			\cmidrule{3-9}
			& \multirow{2}{*}{\rotatebox[origin=c]{90}{\centering F}} 
			&   Binary & 35.39 & 60.55 & 0.5045 & 35.86 & 60.19 & 0.4928 \\
			& & NE & 35.51 & 60.53 & 0.5078 & 36.38 & 60.65 & 0.4871 \\
            \cmidrule{3-9}
            & \multirow{2}{*}{\rotatebox[origin=c]{90}{\centering PL}} 
			&   Binary & 32.49 & 59.17 & 0.5302 & 33.81 & 59.43 & 0.5004 \\ 
			& & NE & 32.94 & 59.58 & 0.522 & 33.49 & 59.35 & 0.501 \\
            \cmidrule{3-9}
            & \multirow{2}{*}{\rotatebox[origin=c]{90}{\centering PG}} 
			&   Binary & 32.29 & 58.88 & 0.5327 & 34.08 & 59.26 & 0.5004 \\
			& & NE & 32.73 & 59.28 & 0.5246 & 33.71 & 59.17 & 0.5011 \\
		\cmidrule{2-9}
		\multirow{2}{*}{\rotatebox[origin=c]{90}{\centering }} 
			& & RBMT & 24.52 & 51.71 & 0.6058 & 23.27 & 49.41 & 0.6288 \\
			& & Google & 38.69 & 61.28 & 0.4942 & 40.08 & 62.07 & 0.4711 \\ 
    \end{tabular}
    \caption{Results for the NE focused models when tested with the generic test set. F refers to the models trained with the NE tag as a word feature, and PL and PG to the model trained to replace NEs with the corresponding token and translating the contents with Lucy LT and Google Translate respectively. Binary models only classify words as NE or not NE, while NE models classifies each NE with the corresponding class according to CoreNLP. All BPE models are statistically significantly better than their word-based counterparts, and all models are statistically significantly better than RBMT. }
    \label{tab:results_ne_ge}
\end{table}

\begin{table}[th!]
    \centering
    \small
    \begin{tabular}{rrr@{\hskip 3\tabcolsep}SSS[table-format=0.4]@{\hskip 3\tabcolsep}SSS[table-format=0.4]}
		\multicolumn{3}{r@{\hskip 3\tabcolsep}}{} & \multicolumn{3}{c@{\hskip 3\tabcolsep}}{English$\rightarrow$Spanish} & \multicolumn{3}{c}{English$\leftarrow$Spanish}\\ 
		\multicolumn{3}{r@{\hskip 3\tabcolsep}}{} & {BLEU\hib}  & {chrF\hib} & {TER\lib} & {BLEU\hib} & {chrF\hib} & {TER\lib} \\ 
		\midrule
		\multirow{10}{*}{\rotatebox[origin=c]{90}{\centering Word}} 
			& & Baseline & 28.97 & 56.42 & 0.5575 & 27.92 & 52.23	 & 0.5731 \\
			\cmidrule{3-9}
			& \multirow{2}{*}{\rotatebox[origin=c]{90}{\centering F}} 
			&   Binary & 28.56	 &  56.24 & 0.5634 & 27.56 & 51.98 & 0.5736 \\
			& & NE & 28.71	 & 56.48 & 0.5662 & 27.47 & 51.88 & 0.5758 \\
			\cmidrule{3-9}
			& \multirow{2}{*}{\rotatebox[origin=c]{90}{\centering PL}} 
			&   Binary & 25.88	 & 55.15 & 0.5851 & 26.27 & 51.79 & 0.6062 \\
			& & NE & 27.07	 & 55.88 & 0.5746 & 25.25 & 51.47 & 0.6138 \\ 
            \cmidrule{3-9}
			& \multirow{2}{*}{\rotatebox[origin=c]{90}{\centering PG}} 
			&   Binary & 26.42 & 56.04 & 0.5767 & 27.37 & 53.09 & 0.5928 \\
			& & NE & 27.64 & 56.79 & 0.5656 & 26.29 & 52.74 & 0.6005 \\ 
		\cmidrule{2-9}
		\multirow{7}{*}{\rotatebox[origin=c]{90}{\centering BPE}} 
			& & Baseline & 30.63 & 59.3 & 0.5508  & 32.11 & 59.77 & 0.5489 \\
			\cmidrule{3-9}
			& \multirow{2}{*}{\rotatebox[origin=c]{90}{\centering F}} 
			&   Binary & 25.78 & 53.04 & 0.6544 & 24.34 & 50.22 & 0.7045 \\
			& & NE & 26.32	 & 53.81 & 0.6521 & 26.51	 & 52.05 & 0.6716 \\
            \cmidrule{3-9}
            & \multirow{2}{*}{\rotatebox[origin=c]{90}{\centering PL}} 
			&   Binary & 26.24	 & 57.18 & 0.5765 & 28.09 & 56.67 & 0.5753 \\
			& & NE & 27.7 & 57.71 & 0.5696 & 27.21 & 56.67 & 0.5845 \\ 
            \cmidrule{3-9}
            & \multirow{2}{*}{\rotatebox[origin=c]{90}{\centering PG}} 
			&   Binary & 26.84 & 58.05 & 0.5676 & 29.18 & 57.94 & 0.5621 \\ 
			& & NE & 28.27 & 58.64 & 0.5608 & 28.17 & 57.94 & 0.5708 \\ 

		\cmidrule{2-9}
		\multirow{2}{*}{\rotatebox[origin=c]{90}{\centering }} 
        	& & RBMT & 22.95 & 52.11 & 0.6255 & 21.24 & 49.38 & 0.6749 \\
    		& & Google & 36.19 & 61.66 & 0.4964 & 43.65 & 66.27 & 0.4458 \\
    \end{tabular}
    \caption{Results for the NE focused models when tested with the NE focused test set. F refers to the models trained with the NE tag as a word feature, and PL and PG to the model trained to replace NEs with the corresponding token and translating the contents with Lucy LT and Google Translate respectively.}
    \label{tab:results_ne_es}
\end{table}

\begin{table}[th!]
    \centering
    \small
    \begin{tabular}{crS[table-format=4.0]S[table-format=4.0]S[table-format=4.0]S[table-format=4.0]}

& & \multicolumn{2}{c}{Generic} & \multicolumn{2}{c}{Specific} \\
& & \multicolumn{1}{c}{EN$\rightarrow$ES} & \multicolumn{1}{c}{EN$\leftarrow$ES} & \multicolumn{1}{c}{EN$\rightarrow$ES} & \multicolumn{1}{c}{EN$\leftarrow$ES} \\ 
\midrule
& Baseline & 	5026 &  4249       &  4465    &  8662  \\
\multirow{2}{*}{\rotatebox[origin=c]{90}{\centering F}} 
& Binary & 	    4915 &  4380       &  4242   &  8995   \\
& NE & 	        4805 &  4331       &  4469    &  9368    \\
\multirow{2}{*}{\rotatebox[origin=c]{90}{\centering P\_}} 
& Binary & 	3020 &  1822       & 2969    &  7252    \\
& NE & 	    2788 &  1648       & 2977    &   7073  \\

\end{tabular}
    \caption{Number of \texttt{<unk>} tokens generated by each approach. Both PG and PL have the same number of \texttt{<unk>}.}
    \label{tab:results_ne_unks}
\end{table}

\tab{results_ne_ge} and \tab{results_ne_es} show the results of the evaluation of the NE-focused models when tested with the generic and specific datasets respectively. Models using the NE feature are not significantly different from the baseline, while the models using the replacement strategy lower the performance of the system. Still, the models using protected sequences reduce the number of \texttt{<unk>} tokens in the hypotheses for the specific evaluation corpus, as shown in \tab{results_ne_unks}. Producing less \texttt{<unk>} may help to improve the adequacy of the sentences.

\subsubsection{Evaluation Focused on Terminology Injection}

\begin{table}[th!]
    \centering
    \small
     \begin{tabular}{rrr@{\hskip 3\tabcolsep}SSS[table-format=0.4]@{\hskip 3\tabcolsep}SSS[table-format=0.4]}
		\multicolumn{3}{r@{\hskip 3\tabcolsep}}{} & \multicolumn{3}{c@{\hskip 3\tabcolsep}}{English$\rightarrow$Spanish} & \multicolumn{3}{c}{English$\leftarrow$Spanish}\\ 
		\multicolumn{3}{r@{\hskip 3\tabcolsep}}{} & {BLEU\hib}  & {chrF\hib} & {TER\lib} & {BLEU\hib} & {chrF\hib} & {TER\lib} \\ 
		\midrule
		\multirow{10}{*}{\rotatebox[origin=c]{90}{\centering Word}} 
		    & & Baseline & 34.22 & 58.87 & 0.5128 &     34.59 & 58.29 & 0.499 \\
			& & EMEA  & 10.97 & 32.96 & 0.8227 & 8.839 & 33.09 & 0.7924 \\
			& & PT & 34.23$\ast$ & 55.88 & 0.5126$\ast$ & 34.6$\ast$ & 55.49 & 0.4988$\ast$ \\
			\cmidrule{3-9}
			& \multirow{2}{*}{\rotatebox[origin=c]{90}{\centering F}} 
			&   MED & 34.6 & 58.65 & 0.5137 &   34.71 & 58.38 & 0.4965 \\
			& & MED+ & 34.23 & 58.74 & 0.5151 &    34.56 & 58.16 & 0.5003 \\
			\cmidrule{3-9}
			& \multirow{2}{*}{\rotatebox[origin=c]{90}{\centering PL}} 
			&   MED & 34.3 & 58.63 & 0.5138 &        34.23 & 58.2 & 0.5001 \\
			& & MED+  & 34.45 & 58.74 & 0.5127 & 34.04 & 58 & 0.5003 \\
            \cmidrule{3-9}
			& \multirow{2}{*}{\rotatebox[origin=c]{90}{\centering PG}} 
			&   MED & 34.34 & 58.68 & 0.5134 &    34.27 & 58.27 & 0.4997 \\
			& & MED+  & 34.48 & 58.78 & 0.5123 & 34.08 & 58.07 & 0.4998 \\
		\cmidrule{2-9}
		\multirow{7}{*}{\rotatebox[origin=c]{90}{\centering BPE}} 
			& & Baseline & 35.68 & 60.38 & 0.5074 &   36.26 & 60.55 & 0.4838 \\
			& & EMEA & 10.64 & 35.54 & 0.8437 &  8.998 & 35.45 & 0.7967 \\
			\cmidrule{3-9}
			& \multirow{2}{*}{\rotatebox[origin=c]{90}{\centering F}} 
			&   MED & 35.76 & 60.49 & 0.5045 &  35.87 & 60.2 & 0.4905 \\ 
			& & MED+ & 35.53 & 60.42 & 0.504 &     36.18 & 60.63 & 0.4857 \\
            \cmidrule{3-9}
            & \multirow{2}{*}{\rotatebox[origin=c]{90}{\centering PL}} 
			&   MED & 35.7 & 60.64 & 0.5033 & 35.65 & 60.28 & 0.4896 \\
			& & MED+  & 35.65 & 60.49 & 0.5038 & 36.03 & 60.41 & 0.4884 \\
            \cmidrule{3-9}
            & \multirow{2}{*}{\rotatebox[origin=c]{90}{\centering PG}} 
			&   MED  & 35.74 & 60.68 & 0.5029 &        35.69 & 60.35 & 0.4891 \\
			& & MED+  & 35.69 & 60.53 & 0.5034 &  36.07 & 60.48 & 0.488 \\

		\cmidrule{2-9}
		\multirow{2}{*}{\rotatebox[origin=c]{90}{\centering }} 
        	& & RBMT & 24.52 & 51.71 & 0.6058 &      23.27 & 49.41 & 0.6288 \\
    		& & Google & 38.69 & 61.28 & 0.4942 &    40.08 & 62.07 & 0.4711 \\
    \end{tabular}
    \caption{Results for the terminology focused models when tested with the generic test set. Models marked with $\ast$ are significantly better than the NMT Baseline model. All BPE models are statistically significantly better than their word-based counterparts, and all models are statistically significantly better than RBMT.}
    \label{tab:results_te_ge}
\end{table}

\begin{table}[th!]
    \centering
    \small
    \begin{tabular}{rrr@{\hskip 3\tabcolsep}SSS[table-format=0.4]@{\hskip 3\tabcolsep}SSS[table-format=0.4]}
		\multicolumn{3}{r@{\hskip 3\tabcolsep}}{} & \multicolumn{3}{c@{\hskip 3\tabcolsep}}{English$\rightarrow$Spanish} & \multicolumn{3}{c}{English$\leftarrow$Spanish}\\ 
		\multicolumn{3}{r@{\hskip 3\tabcolsep}}{} & {BLEU\hib}  & {chrF\hib} & {TER\lib} & {BLEU\hib} & {chrF\hib} & {TER\lib} \\ 
		\midrule
		\multirow{11}{*}{\rotatebox[origin=c]{90}{\centering Word}} 
			& & Baseline & 27.46 & 51.37 & 0.5904 & 28.24 & 49.06 & 0.5972 \\
			& & EMEA & 58.64$\ast$ & 76.29$\ast$ & 0.3536$\ast$ & 63.91$\ast$ & 78.13$\ast$ & 0.3178$\ast$ \\
            & & BT$_1$ & 38.67$\ast$ & 61.06$\ast$ & 0.493$\ast$ & 37.93$\ast$ & 57.02$\ast$ & 0.5122$\ast$ \\
            & & BT$_2$ & 33.24$\ast$ & 57.78$\ast$ & 0.53$\ast$ & 33.94$\ast$ & 54.52$\ast$ & 0.5432$\ast$ \\
            & & PT & 32.82$\ast$ & 59.43$\ast$ & 0.5475$\ast$ & 34.38$\ast$ & 56.92$\ast$ & 0.5512$\ast$ \\
			\cmidrule{3-9}
			& \multirow{2}{*}{\rotatebox[origin=c]{90}{\centering F}} 
			&   MED & 27.56 & 51.23 & 0.588 & 28.06 & 48.72 & 0.6009 \\
			& & MED+ & 27.76 & 51.7 & 0.5903 & 27.96 & 48.59 & 0.6026 \\
			\cmidrule{3-9}
			& \multirow{2}{*}{\rotatebox[origin=c]{90}{\centering PL}} 
			&   MED & 27.56 & 51.72 & 0.5938 & 27.78 & 48.71 & 0.6046 \\
			& & MED+ & 27.62 & 51.81 & 0.5956 & 27.1 & 48.89 & 0.6019 \\ 
            \cmidrule{3-9}
			& \multirow{2}{*}{\rotatebox[origin=c]{90}{\centering PG}} 
			&   MED & 27.7 & 51.86 & 0.5922 & 28.4 & 49.35 & 0.5993 \\
			& & MED+ & 27.75 & 51.95 & 0.594 & 27.73 & 49.54 & 0.5965 \\ 
		\cmidrule{2-9}
		\multirow{11}{*}{\rotatebox[origin=c]{90}{\centering BPE}} 
			& & Baseline & 32.87 & 59.52 & 0.563  & 33.1 & 57.39 & 0.566 \\
			& & EMEA & 58.56${\ast}$ & 76.37${\ast}$ & 0.3564${\ast}$ & 63.52${\ast}$ & 78.4${\ast}$ & 0.323${\ast}$ \\
            & & BT$_1$ & 42.68${\ast}$ & 66.28${\ast}$ & 0.4704${\ast}$ & 44.18${\ast}$ & 64.56${\ast}$ & 0.4645${\ast}$ \\
            & & BT$_2$ & 45.22${\ast}$ & 68.02${\ast}$ & 0.4501${\ast}$ & 46.71${\ast}$ & 66.51${\ast}$ & 0.445${\ast}$ \\
            & & BT$_3$ & 46.26${\ast}$ & 68.59${\ast}$ & 0.4387${\ast}$ & 47.7${\ast}$ & 67.06${\ast}$ & 0.4377${\ast}$ \\
			\cmidrule{3-9}
			& \multirow{2}{*}{\rotatebox[origin=c]{90}{\centering F}} 
			&   MED & 32.45 & 59.45 & 0.5591 & 32.45 & 56.85 & 0.5679 \\
			& & MED+ & 31.21 & 58.2 & 0.5739 & 33.21 & 57.37 & 0.5648 \\
            \cmidrule{3-9}
            & \multirow{2}{*}{\rotatebox[origin=c]{90}{\centering PL}} 
			&   MED & 31.23 & 58.85 & 0.573 & 31.62 & 56.43 & 0.5766 \\
			& & MED+ & 31.66 & 59.23 & 0.5702 & 33.08 & 57.48 & 0.5591 \\ 
            \cmidrule{3-9}
            & \multirow{2}{*}{\rotatebox[origin=c]{90}{\centering PG}} 
			&   MED & 31.38 & 59.02 & 0.5714 & 32.26 & 57.08 & 0.5714 \\ 
			& & MED+ & 31.82 & 59.39 & 0.5687 & 33.72 & 58.12 & 0.5537 \\ 

		\cmidrule{2-9}
		\multirow{2}{*}{\rotatebox[origin=c]{90}{\centering }} 
        	& & RBMT & 27.44 & 55.68 & 0.6021 & 27.97 & 53.63 & 0.63 \\
    		& & Google & 46.82 & 68.71 & 0.4378 & 47.52 & 67.63 & 0.4434 \\
    \end{tabular}
    \caption{Results for the terminology focused models when tested with the terminology focused test set. PT uses the OpenNMT phrase table feature with the dictionary extracted from Lucy LT, BT$_i$ refers to the model trained on the back-translated corpus on the $i$-th iteration, F refers to the models trained with the \texttt{MED} tag as a word feature, and PL and PG to the model trained replacing terms with the \texttt{MED} token and translating the contents with Lucy LT and Google Translate respectively. {MED+} models duplicate those lines that have terms, leaving them untouched, while processing the other. Models marked with $\ast$ are significantly better than the NMT Baseline model, and all BPE models are statistically significantly better than RBMT.}
    \label{tab:results_te_es}
\end{table}

Results of the automatic evaluation for terminology injection can be seen in \tab{results_te_ge} (with the generic test set) and \tab{results_te_es} (with the specific EMEA test set). EMEA was trained with the corpus labelled English-Spanish (EMEA) in \tab{data_stats}, and should be treated as an upper bound. During our experiments, we observed that the back-translated models outperform all other alternatives. While the BPE-level model improved in the first three back-translation rounds, the word-level model only improved on the first back-translation round. Still, back-translation is not only injecting terminology into the models but also other linguistic information, as full sentences are being fed to the system. 

All the approaches that add information to the word-level models performed slightly better than the baseline model, but the opposite happened for the BPE models. Only \num{9759} sentences in the train set had medical terms, limiting the effect of this approach. The mode using the phrase table replacement is replacing the \texttt{<unk>} tokens with information contained in the provided phrase table, hence obtaining a very small but significant improvement over the baseline in the generic test set, but a major one on the specific test set.

\subsection{Qualitative Evaluation}

In this section, we describe the qualitative evaluation of the different models.

\subsubsection{General Evaluation}

\tab{example} analyses a sentence translated using all different models from Spanish to English. The analysis showed that, even when RBMT makes some grammatical mistakes, the sentence still conveyed the correct message. Nevertheless, it was the only hypothesis with a BLEU of \num{0}, as it shared no four-gram with the reference, and was the hypothesis with the highest TER. The baseline model hypothesis was tied for the best TER score and the second best BLEU score, but it failed to convey the proper message, as it lacked translation for \textit{easing of price increases}. 

\begin{table}[th!]
    \centering
    \small
    \begin{tabular}{p{0.09\linewidth} p{0.67\linewidth} S[table-format=2.2] S[table-format=0.2]}
    \toprule
Source & {Pese a que los incrementos de los precios fueron menores en el segundo semestre de 2008 , los precios siguen siendo muy elevados .} & {BLEU} & {TER}\\
Reference &     Despite an easing of price increases in the second half of 2008, prices remain at very high levels. &  \\
\midrule
Baseline &      Despite the increases in prices in the second half of 2008, prices remain very high. & 47.48 & 0.35 \\
CAT-CL &		\textit{Although} price increases were \textbf{minor} in the second half of 2008, prices remain very high. & 47.48 & 0.35 \\
CAT-CL D &		\textit{Although} increases in prices were lower in the second half of 2008, prices remain high. & 44.50 & 0.45\\
POS &		\textit{Despite the fact that} price increases were lower in the second half of 2008, prices remain very high. & 48.25 & 0.35\\
CAT-CL L &		\textit{Although} price increases were lower in the second half of 2008, prices remain very high. & 47.48 & 0.35\\
Tree &		\textit{Although} \textbf{prices of} prices were lower in the second half of 2008  prices remain very high. & 45.51 & 0.40\\
RBMT &		\textit{Even though} the increases of the prices \textbf{were smaller} in the second \textit{semester} of 2008, the prices \textbf{keep being} \textit{sky-high}. & 0.00 & 0.70\\
Google &		\textit{Although} the price increases \textit{were lower} in the second half of 2008, prices \textit{are still} very high. & 41.81 & 0.40\\
    \bottomrule
    \end{tabular}
    \caption{Qualitative analysis of a sentence translated by all models for Spanish to English translation. Fragments in bold face are translation mistakes, and fragments in italics are translation alternatives that, while being penalised by TER and BLEU, can be considered correct.}
    \label{tab:example} 
\end{table}

\subsubsection{Evaluation Focused on Morphological Information}

\begin{table}[th!]
    \centering
    \small
    \begin{tabularx}{\linewidth}{p{0.1\linewidth} X S[table-format=2.2] S[table-format=0.2]}
    \toprule
Source & {He could have been sent home tomorrow if only you \textbf{had kept} quiet.} & {chrF} & {TER}\\

Reference & Hubiera podido salir mañana si no hubiera abierto la boca.  &  \\
\midrule
Baseline & Podría haber sido enviado a casa mañana si sólo \textbf{se hubiera mantenido} callado. & 36.3705 & 1.0 \\
BPE & Podría haber sido enviado a casa mañana si \textbf{hubieras mantenido} silencio. & 36.0863 & 0.9090909090909091 \\
CAT-CL & Podría haber sido enviado a casa mañana si sólo \textbf{hubieras mantenido} silencio.
  & 37.7586 &1.0\\
CAT-CL D & Podría haber sido enviado a casa mañana si sólo \textbf{te quedaba} callado.
  & 28.1555 & 1.0\\
POS & Él podría haber sido enviado a casa mañana si sólo \textbf{se hubiera} callado.  & 35.4873 & 1.0909090909090908\\
CAT-CL L &	Podría haber sido enviado a casa mañana si sólo \textbf{hubieras mantenido} callado.
 & 37.8267 & 1.0\\
Tree & Podría haber sido enviado a casa mañana si sólo \textbf{hubieras mantenido} silencio. & 37.7586 & 1.0\\
RBMT &	Se le podría haber enviado a casa mañana si solamente usted \textbf{hubiera estado} callado. & 35.4391 & 1.0909090909090908\\
Google & Podría haber sido enviado a casa mañana si solo \textbf{hubieras guardado} silencio. & 38.2189 & 1.0\\
    \bottomrule
    \end{tabularx}
    \caption{Qualitative analysis of morphology. Most BLEU scores were 0; instead, chrF was used. }
    \label{tab:morphoexample} 
\end{table}

\tab{morphoexample} shows an example of a sentence translated with all the different models. Even when translating a fairly complex sentence that accepts many different options, most of the models are able to produce tenses that keep the sense of the sentence. The reference translation is fairly idiomatic, thus no model perfectly matches with it.

\subsubsection{Evaluation Focused on Named Entities}

\begin{table}[th!]
    \centering
    \small
    \begin{tabularx}{\linewidth}{p{0.1\linewidth} X S[table-format=2.2] S[table-format=0.2]}
    \toprule
Source & \textbf{Langdon} deduce que \textbf{Sauniere} fue miembro del \textbf{priorato} de \textbf{Sion}, una sociedad secreta asociada a la \textbf{orden del temple}. & {chrF} & {TER} \\

Reference & \textbf{Langdon} deduces from this that \textbf{Sauniere} was a member of the \textbf{priory} of \textbf{Sion}, a secret society associated with the \textbf{knights templar}. & & \\
\midrule
Baseline & \texttt\textbf{<unk>} follows that \texttt\textbf{<unk>} was a member of the \textbf{priory} of \texttt\textbf{Zion}, a secret society associated with the \textbf{order of the temple}. & 59.0566 & 0.4166666666666667 \\
BPE & \textbf{Langdon} deduces that \textbf{Sapuniere} was a member of the \textbf{priory}, a secret society associated with the \textbf{order of the temple}. & 70.8927 & 0.3333333333333333 \\
F & \textbf{Langdon} argues that \textbf{Sauniere} was a member of the \textbf{priory} of \textbf{Zion}, a secret society associated with the \textbf{order of the temple}. & 72.3719 & 0.3333333333333333 \\
P\_ & \texttt{PERSON} states that \texttt{PERSON} was a member of the \texttt{ORGANIZATION} of \texttt{MISC}, a secret society associated with the \texttt{ORGANIZATION} & {--} & {--} \\
PL & \textbf{Langdon} states that \textbf{Sauniere} was a member of the \textbf{priorate} of \textbf{Zion}, a secret society associated with the \textbf{order of the temper}. & 69.2095 & 0.375 \\
PG &  \textbf{Langdon} states that \textbf{Sauniere} was a member of the \textbf{priory} of \textbf{Zion}, a secret society associated with the \textbf{order of the temple}. & 72.3719 & 0.3333333333333333 \\
RBMT & \textbf{Langdon} deduces that \textbf{Sauniere} was a member of the \textbf{priorate} of \textbf{Zion}, a secret society associated to the \textbf{order to the temper}. & 68.8098 & 0.375 \\
Google & \textbf{Langdon} deduces that \textbf{Sauniere} was a member of the \textbf{priory} of \textbf{Sion}, a secret society associated with the \textbf{order of the temple}. & 80.7856 & 0.25 \\
    \bottomrule
\end{tabularx}
    \caption{Qualitative analysis of named entities. F refers to the BPE-level model with NE classes marked with features, P\_ refers to the BPE-level model with NE replaced with a unique token for each NE class, then translated using Lucy LT (PL) or Google Translate (PG). Finally, RBMT refers to Lucy LT translation and Google to Google Translate. Most BLEU scores were 0; instead, chrF was used.}
    \label{tab:neexample} 
\end{table}

The example sentence in \tab{neexample} shows that using the proposed approach can lead to improved translations. Langdon and Sauniere are both unknown words in the Baseline and BPE cases, leading to \texttt{<unk>} tokens in the output of the former, and the incorrect translation \textit{Sapuniere} for the latter. No model was able to properly translate \textit{orden del temple} to \textit{knights templar}; in some cases, the word \textit{temple} is untranslated (such as in PG or Google), and in the others it gets improperly translated to \textit{temper}.

\subsubsection{Evaluation Focused on Terminology Injection}


The example sentence in \tab{termexample} shows that using this approach to focus on terminology can lead to better translations. The Spanish word \textit{vaso} can be translated to \textit{glass} or \textit{cup} (e.g. a glass of water) or to 
\textit{vessel} or \textit{vein} (e.g. a blood vessel), depending upon the context of the input sentence. The example sentence in \tab{termexample} shows that both baseline models use the first sense when translating, but when replacing the identified term \textit{vaso sanguíneo} with the tag \texttt{MED}, the correct sense is used. As a side effect, the produced hypothesis keep the passive voice characteristic of medical text. Still, no model is able to properly translate \textit{atravesar}; all models use the incorrect sense of going through or crossing, instead the proper term, entering. 

We also analysed how Lucy LT and Google Translate certain medical terminology in \tab{medterm_size}. Translations of medical terms produced by Google Translate appear to have a higher overlap withe the reference corpus, hence leading to higher evaluation score.

\begin{table}[th!]
    \centering
    \small
    \begin{tabularx}{\linewidth}{p{0.1\linewidth} X S[table-format=2.2] S[table-format=0.2]}
    \toprule
Source &  Debe tenerse precaución para no atravesar un \textbf{vaso sanguíneo}. & {chrF} & {TER} \\
Reference &  Care should be taken to ensure that a \textbf{blood vessel} has not been entered. & & \\
\midrule
Baseline &  You must be careful not to go through a \textbf{glass of blood}. & 18.21 & 0.7333\\
BPE &  You must be careful to not go through a \textbf{blood glass}. & 19.51 & 0.6666 \\
F & You must be careful not to get through a \textbf{blood cup}. & 18.67 & 0.6666 \\
P\_ &  Caution must be taken not to cross a \texttt{MED}. & {--} & {--} \\
PL &  Caution must be taken not to cross a \textbf{blood vessel}. & 32.75 & 0.5333 \\
PG &  Caution must be taken not to cross a \textbf{blood vessel}. & 32.75 & 0.5333 \\
RBMT &  Precaution must be had not to go across a \textbf{blood vessel}. & 27.64 & 0.6 \\
Google & Care must be taken not to cross a \textbf{blood vessel}. & 34.15 & 0.4666 \\
    \bottomrule
\end{tabularx}
    \caption{Qualitative analysis of terminology injection. F refers to the BPE-level model with terms marked with features, P\_ refers to the BPE-level model with terms replaced with the \texttt{MED} token, then translated using Lucy LT (PL) or Google Translate (PG). Finally, RBMT refers to Lucy LT translation and Google to Google Translate. Most BLEU scores were 0; instead, chrF was used.}
    \label{tab:termexample} 
\end{table}

\begin{table}[th!]
    \centering
    \small
    \begin{tabularx}{\linewidth}{XSXSXS}
        \toprule
        \multicolumn{2}{c}{Spanish} & \multicolumn{2}{c}{RBMT} & \multicolumn{2}{c}{Google}  \\
         \midrule
         dosis & 828 & shot & 1 & dose & 645 \\
         medicamento & 267 & medication & 8 & medicine & 236 \\
         frasco & 44 & vial & 2 & bottle & 36 \\
         análisis & 42 & test & 12 & analysis & 18 \\
         presión arterial & 36 & arterial tension & 0 & blood pressure & 29 \\
         miocardio & 23 & coronary & 3 & myocardial & 17 \\
         ictericia & 6 & icterus & 0 & jaundice & 4 \\
        \bottomrule
    \end{tabularx}
    \caption{Terminology selection for each MT system. The number indicates how many times the word appears in the Spanish side of the test set, and how many times the proposed translation appeared in the corresponding reference.}
    \label{tab:medterm_size}
\end{table}

\section{Conclusions and Future Work}

In this work, we explored the use of rule-based machine translation (RBMT) knowledge to improve the performance of neural machine translation (NMT) models in an under-resourced scenario, showing that the models had limited ability to learn from the external information. 

We also tested different approaches to inject named entities (NE) and terminological expressions contained in the RBMT model to NMT. The approaches treat the NMT model as a black box, that is, in such a way that there is no need to know or modify the inner workings of the system, thus being applicable to any model, implementation and architecture. Only the approaches injecting terminology in word-based models improved the baseline, albeit not statistically significantly. In some scenarios, the use of some approaches led to translations that, while not having a significantly different automatic evaluation score, appear to be closer to the style of the targeted text; namely, in the case of terminology translation, some strategies managed to retain the passive voice of the corpus. 

One of the paths of our future work will further focus on the extraction of RBMT knowledge and the inclusion of transfer rules to improve the performance of the NMT model. The model that was trained following the structure with the parse tree was not able to properly deal with the information, and generally performed worse than the rest; integrating this information differently might produce better results. 

A second path is using approaches that modify the architecture of the neural network. For example, using multiple encoders to take both the source sentence and the output of the RBMT system. This approach has been used to improve the performance of NMT~\cite{N16-1004}. As previously mentioned, corpus-based MT gives limited control over the output to the user, especially when dealing with homographs and terminology; instead, RBMT gives total control. Combining the source sentence with the RBMT output that contains the user-selected translations might lead to improvements in domain-specific or low resource scenarios.


Finally, we also plan to leverage information contained in other freely available RBMT systems, such as Apertium, that contains features similar to the ones used in this work.


\section*{Acknowledgments}
This publication has emanated from research supported in part by a research grant from Science Foundation Ireland (SFI) under Grant Number SFI/12/RC/2289, co-funded by the European Regional Development Fund, and the Enterprise Ireland (EI) Innovation Partnership Programme under grant agreement No IP20180729, NURS - Neural Machine Translation for Under-Resourced Scenarios.


\bibliographystyle{spbasic}
\bibliography{biblio}

\newpage

\section*{Appendix I: Models}

\begin{itemize}
    \item \textbf{Baseline:} no extra features or protected sequences
    \item \textbf{CAT-CL:} category (\texttt{CAT}) and class (\texttt{CL}) ambiguity classes as features
    \item \textbf{CAT-CL D:} category (\texttt{CAT}) and class (\texttt{CL}) disambiguated with Lucy LT as features
    \item \textbf{CAT-CL L:} category (\texttt{CAT}) and class (\texttt{CL}) disambiguated with CoreNLP as features
    \item \textbf{Tree:} category (\texttt{CAT}) and class (\texttt{CL}) from Lucy LT as features, extra tokens for tree structure
    \item \textbf{POS:} CoreNLP POS tags as features
    \item \textbf{RBMT:} Lucy LT translation
    \item \textbf{Google:} Google Translate translation
    \item \textbf{BT$_{i}$:} Back-translated corpus on the $i$-th iteration
    \item \textbf{MED:} Corpus with medical (\texttt{MED}) domain terms tagged
    \item \textbf{MED+:} Corpus with medical (\texttt{MED}) domain terms tagged, sentences with \texttt{MED} domain get duplicated and tagged as generic (\texttt{GEN})
    \item \textbf{F:} Terms or named entities tagged as word features
    \item \textbf{PL:} Terms or named entities replaced by tag, content translated with Lucy LT
    \item \textbf{PG:} Terms or named entities replaced by tag, content translated with Google Translate
\end{itemize}

\end{document}